\documentclass[12pt]{article}

\usepackage{sbc-template}
\usepackage{graphicx,url}
\usepackage{subcaption}
\usepackage[utf8]{inputenc}
\usepackage[english]{babel}
\usepackage{amsmath}
\usepackage{multirow}
\usepackage{booktabs}
\usepackage{float}
\title{Aycromo: An Open-Source Platform for Automatic Chromosome Detection in Metaphase Images Based on \textit{Deep Learning}}

\author{Jorge L. A. Lima\inst{1}, Filipe R. Cordeiro\inst{1}}

\address{
  Visual Computing Lab, Department of Computing, \\ Universidade Federal Rural de Pernambuco (UFRPE), Brazil
  \email{jorgelucasalima@gmail.com, filipe.rolim@ufrpe.br } 
}

\begin{document}

\maketitle

\begin{abstract}
Chromosome analysis is a fundamental step in the diagnosis of genetic diseases, but the manual karyotyping workflow is time-consuming and heavily dependent on expert specialists, often requiring several days per patient. Although \textit{Deep Learning} models have achieved high performance in chromosome detection, most proposed solutions remain restricted to research prototypes or lack graphical interfaces suitable for clinical use. In this work, we present \textbf{Aycromo}, an open-source desktop platform for AI-assisted cytogenetic analysis. Built on Electron and ONNX Runtime, the tool allows cytogeneticists to load pre-trained models, compare architectures through an integrated benchmarking module, and manually correct detections via an interactive annotation interface, all without command-line interaction. Preliminary experiments on metaphase images from the CRCN-NE dataset demonstrate that YOLOv11 achieves 99.40\% mAP@50, while the platform reduces per-slide analysis to seconds. The code is available at https://github.com/jorgelucasalima/aycromo-electron-front.
\end{abstract}

\section{Introduction}

Chromosome analysis plays a central role in the diagnosis of various medical conditions, including congenital genetic syndromes, such as Down syndrome, and different types of hematological neoplasms \cite{nussbaum2015}. The diagnostic process involves obtaining and analyzing metaphase images, in which chromosomes are identified, counted, and classified into homologous pairs. However, this procedure is entirely manual in conventional clinical practice, requiring high specialization and possibly demanding up to five working days per patient \cite{damatta2013}. This operational bottleneck represents a significant cost in both time and specialized human resources.

With the advancement of \textit{Deep Learning}, models based on convolutional neural networks have started to demonstrate promising performance in detection and classification tasks on medical images. In particular, object detectors from the YOLO (\textit{You Only Look Once}) family \cite{redmon2016yolo} have stood out due to their computational efficiency and high accuracy in several computer vision applications, including automatic chromosome detection \cite{cordeiro25}. However, despite the algorithmic advances, a critical gap persists between the technical performance of the models and their actual clinical applicability: most solutions proposed in the literature do not provide adequate graphical user interfaces (GUI) and operate in a private domain, making them inaccessible to cytogenetics professionals without programming skills \cite{al-kharraz2020}.

In this context, this work presents \textbf{Aycromo}, an open-source desktop platform designed to democratize access to chromosome detection models based on \textit{Deep Learning}. The platform integrates pre-trained models in ONNX format, runs comparative experiments between architectures, and allows manual correction of detections through an intuitive graphical interface. The main contributions of this work are: (i) the development of Aycromo, a chromosome analysis and detection tool with a GUI accessible to specialists without programming knowledge; (ii) an integrated \textit{benchmarking} module for model comparison on the laboratory's own dataset; and (iii) the evaluation of three detection architectures on chromosome images from the CRCN-NE dataset~\cite{cordeiro25}.

\section{Related Work}

The automation of cytogenetic analysis has evolved from classical approaches based on thresholding and mathematical morphology to data-driven methods using \textit{Deep Learning} \cite{xiao2020deepacev2}. Several architectures have been proposed for chromosome classification, including siamese networks \cite{swati2017siamese} and \textit{Varifocal-Net} \cite{qin2019varifocal}, aimed at extracting discriminative regional features between homologous pairs. In the object detection domain, models are divided between two-stage detectors, such as Faster R-CNN \cite{ren2015faster}, widely used in metaphase analysis \cite{xiao2020deepacev2}, and single-stage detectors, such as the YOLO family \cite{redmon2016yolo} and RetinaNet \cite{lin2017focal}. The latter employs the \textit{Focal Loss} to mitigate the imbalance between background regions and objects of interest in the images.

Despite the significant results achieved by these architectures, the literature reveals an important gap regarding the clinical accessibility of these tools. Table~\ref{tab:comparacao_ferramentas} summarizes the main solutions identified, comparing them according to the availability of a graphical interface, code openness, and support for user-driven model extension. The system by Al-Kharraz et al.~\cite{al-kharraz2020} is one of the few exceptions that offers a graphical interface; however, it employs YOLOv2 and remains in the private domain, without allowing the integration of new models. More recent works, such as Kuo et al.~\cite{Kuo2024}, use advanced architectures (Res2Net/BiFPN), but also do not provide a GUI, public code, or extension mechanism. Aycromo emerges to fill this gap by combining state-of-the-art detection with an open platform equipped with a complete graphical interface and native support for the integration of external models via the ONNX format.

\begin{table}[ht]
\centering
\caption{Comparison of chromosome detection tools. \textit{GUI}: graphical user interface; \textit{Domain}: code openness; \textit{Model Ext.}: support for external model integration.}
\label{tab:comparacao_ferramentas}
\scriptsize
\begin{tabular}{lcccc}
\toprule
\textbf{Method} & \textbf{Architecture} & \textbf{GUI} & \textbf{Domain} & \textbf{Model Ext.} \\
\midrule
Xiao et al.~\cite{xiao2020deepacev2}    & Faster R-CNN  & No  & Private & No \\
Al-Kharraz et al.~\cite{al-kharraz2020} & YOLOv2        & Yes & Private & No \\
Kuo et al.~\cite{Kuo2024}               & Res2Net/BiFPN & No  & Private & No \\
Atencia-Jiménez (2024)                  & YOLOv8x       & No  & Public  & No \\
\midrule
\textbf{Aycromo (proposed)} & \textbf{YOLOv11, Faster R-CNN, RetinaNet} & \textbf{Yes} & \textbf{Public} & \textbf{Yes} \\
\bottomrule
\end{tabular}
\end{table}

\section{The Aycromo Platform}

\subsection{Overview}

\begin{figure}[ht]
\centering
\includegraphics[width=0.92\linewidth]{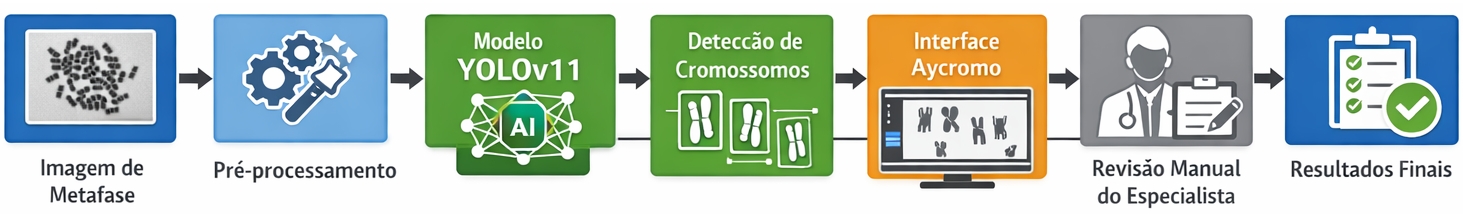}
\caption{Aycromo \textit{pipeline}. The metaphase image is submitted to the selected model and the predictions are displayed for review and manual correction by the specialist.}
\label{fig:pipeline}
\end{figure}

Figure~\ref{fig:pipeline} illustrates the platform's workflow. The user loads a metaphase image, selects the pre-trained model in ONNX format, and triggers the automated detection. The generated bounding boxes are overlaid on the image and displayed in the interface, where the specialist can review, correct, or complement them through the interactive annotation module. Additionally, the \textit{benchmarking} module allows comparing performance metrics between models loaded into the platform, assisting the laboratory in choosing the most suitable architecture for its context.

\subsection{Main Features}

\begin{figure}[ht]
\centering
\begin{subfigure}[t]{0.32\linewidth}
  \includegraphics[width=\linewidth]{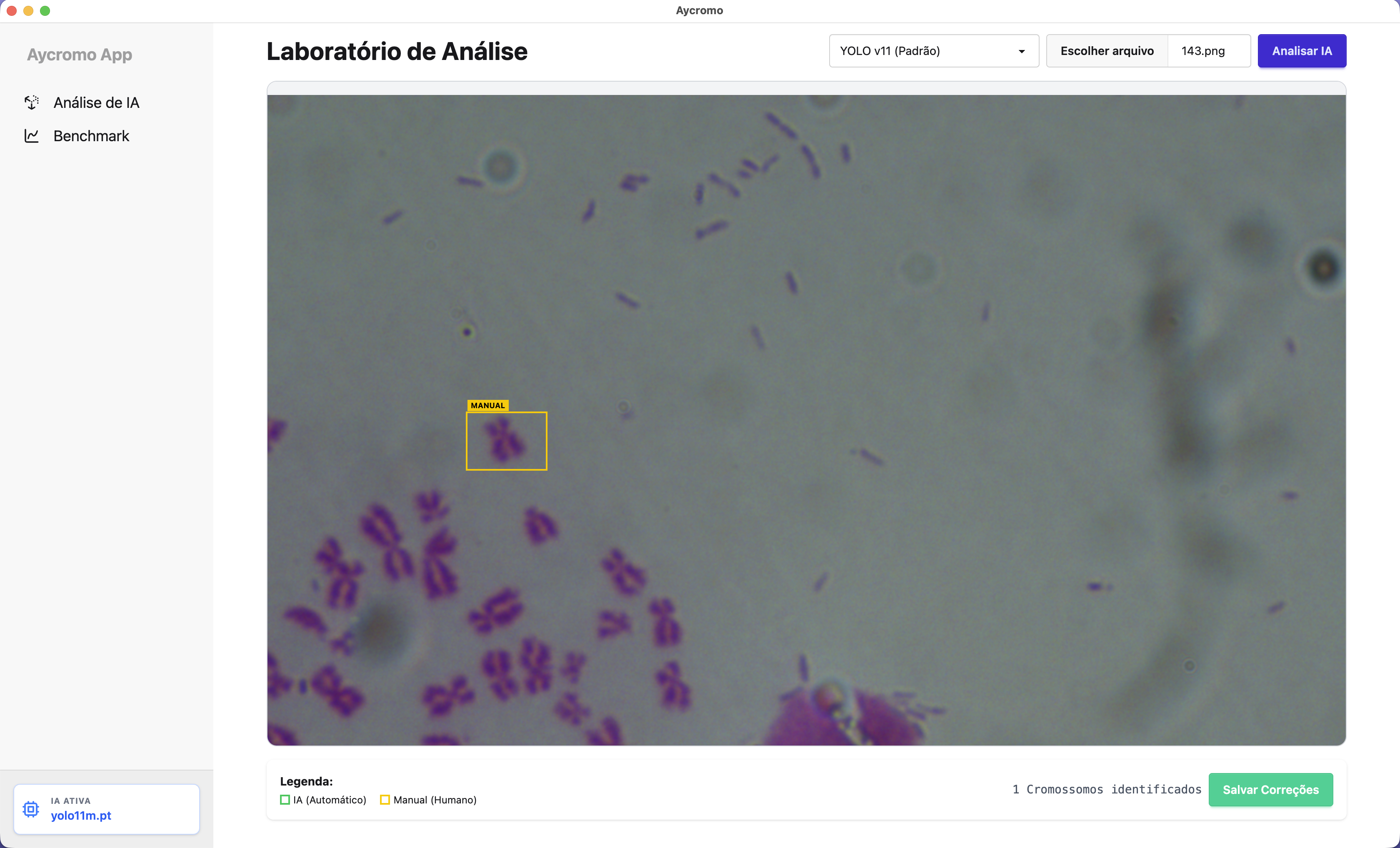}
  \caption{Interactive annotation.}
  \label{fig:anotacao}
\end{subfigure}
\hfill
\begin{subfigure}[t]{0.32\linewidth}
  \includegraphics[width=\linewidth]{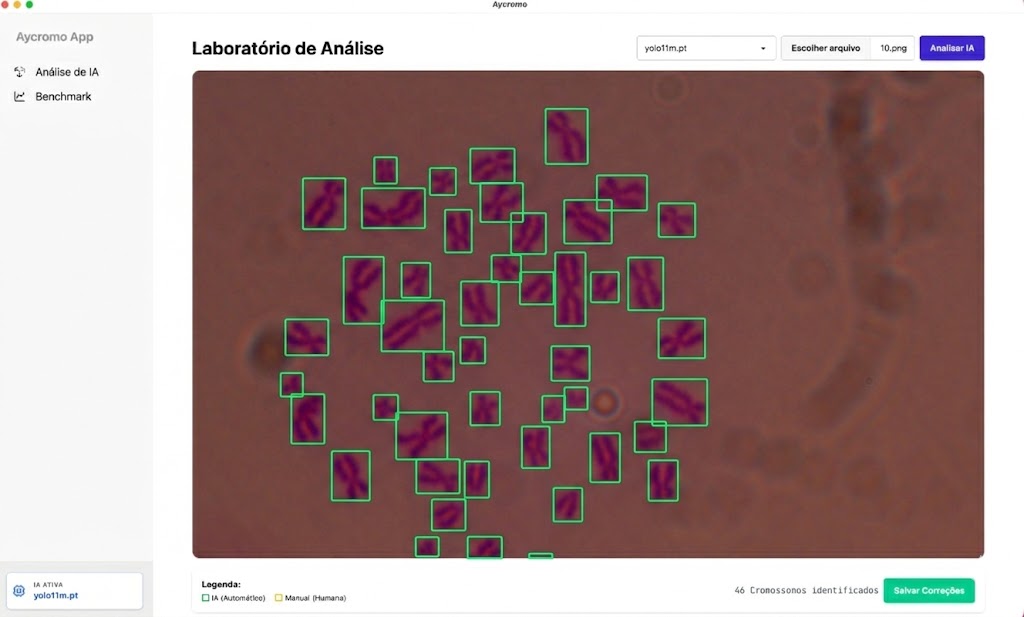}
  \caption{Automatic detection.}
  \label{fig:detecao}
\end{subfigure}
\hfill
\begin{subfigure}[t]{0.32\linewidth}
  \includegraphics[width=\linewidth]{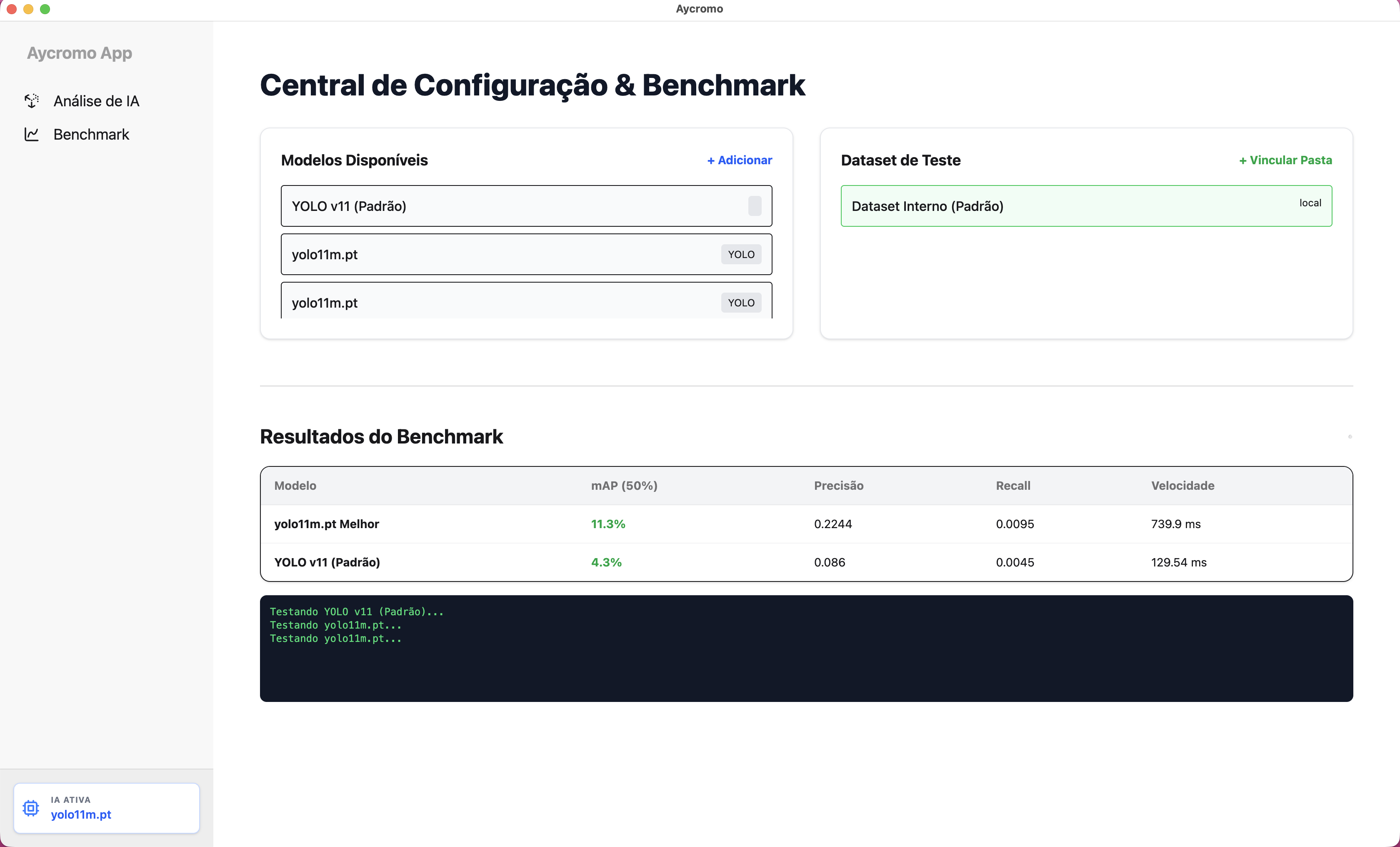}
  \caption{\textit{Benchmarking}.}
  \label{fig:benchmark}
\end{subfigure}
\caption{Aycromo modules: (a) interactive annotation, where the specialist adjusts bounding boxes directly on the metaphase image; (b) automatic detection with confidence indicators; (c) comparative panel of mAP and \textit{Loss} metrics across architectures.}
\label{fig:telas}
\end{figure}

The platform is organized into four complementary modules. The \textbf{model management module} allows loading weights of pre-trained neural networks in ONNX format and switching between them without restarting the application, providing flexibility so that each laboratory can use the model best suited to its imaging conditions. The \textbf{interactive annotation module} (Figure~\ref{fig:telas}, left) provides an interface in which the specialist can add, remove, or adjust bounding boxes directly on the metaphase image. The \textbf{automatic detection module} (Figure~\ref{fig:telas}, center) performs inference from an input image and displays each detected chromosome with its bounding box and confidence indicator. Finally, the \textbf{\textit{benchmarking} module} (Figure~\ref{fig:telas}, right) presents comparative charts of mAP and \textit{Loss} metrics across the evaluated models, allowing the cytogeneticist to validate and compare architectures on their own dataset.

\subsection{Technical Architecture}

Aycromo was developed as a cross-platform desktop application using the Electron \textit{framework} on top of Node.js, allowing it to run on Windows, macOS, and Linux. The graphical interface was built with React and styled with Tailwind CSS, prioritizing a visual experience adapted to the clinical environment. The inference core uses ONNX Runtime \cite{onnx2024}, allowing models trained in different \textit{frameworks}, such as PyTorch or TensorFlow, to be executed directly on local hardware, preserving the privacy of medical data.

\section{Experimental Methodology}

\subsection{Dataset}

The experiments were conducted using the metaphase image dataset from the Centro Regional de Ciências Nucleares do Nordeste (CRCN-NE) \cite{cordeiro25}, consisting of 519 images at $2048\times1536$ pixel resolution, collected in a real clinical environment and manually annotated by specialists. The images present variations in illumination, different levels of chromosomal condensation, and cases of overlapping chromosomes, reflecting the challenges encountered in routine laboratory practice.

\subsection{Evaluated Architectures and Training Protocol}

Three detection architectures were implemented and compared: \textbf{YOLOv11} \cite{ultralytics_yolo11}, \textbf{RetinaNet} \cite{lin2017focal}, and \textbf{Faster R-CNN} \cite{ren2015faster}. YOLOv11 treats detection as a direct regression problem over the entire image, making it suitable for real-time applications. RetinaNet employs \textit{Focal Loss} to focus learning on hard background regions. Faster R-CNN uses a region proposal network to generate candidates before the final classification. The dataset was partitioned into 70\% for training, 15\% for validation, and 15\% for testing. Training employed the SGD optimizer with monitoring of loss functions to avoid \textit{overfitting}. The evaluation metric was mAP@50 (\textit{Mean Average Precision} with an IoU threshold of 0.5), which quantifies the overlap between predicted bounding boxes and reference annotations.

\section{Results and Discussion}

\subsection{Architecture Performance}

Table~\ref{tab:resultados} presents the quantitative results obtained by the evaluated architectures. YOLOv11 achieved the best performance, with 99.40\% mAP@50. Faster R-CNN obtained 97.90\% mAP, while RetinaNet reached 96.21\%. These results are aligned with those of Cordeiro et al.~\cite{cordeiro25}, who also reported superior performance of YOLOv11 in chromosome detection on CRCN-NE images.

\begin{table}[ht]
\centering
\caption{Quantitative performance of the evaluated architectures on the CRCN-NE dataset.}
\label{tab:resultados}
\small
\begin{tabular}{lc}
\toprule
\textbf{Architecture} & \textbf{mAP@50 (\%)} \\
\midrule
YOLOv11 \cite{ultralytics_yolo11} & \textbf{99.40} \\
Faster R-CNN \cite{ren2015faster}  & 97.90          \\
RetinaNet \cite{lin2017focal}      & 96.21          \\
\bottomrule
\end{tabular}
\end{table}

\subsection{Clinical Impact and Usability}

Beyond the quantitative performance, the main contribution of Aycromo is the elimination of the usability barrier that hinders the clinical adoption of these technologies. As evidenced in Table~\ref{tab:comparacao_ferramentas}, Aycromo is the only solution identified that simultaneously combines state-of-the-art detection (YOLOv11), public code availability, a complete graphical interface, and support for user-driven model extension. While conventional manual analysis can take up to five days per patient \cite{damatta2013}, automated triage via Aycromo reduces this time to seconds per slide. The interactive annotation module ensures that the specialist remains in the decision loop, being able to correct errors that typically occur between chromosomal pairs with similar morphology or in regions of overlap. The flexibility to load different ONNX models allows each laboratory to adopt the architecture best suited to its context.

\section{Conclusion}

This work presented Aycromo, an open-source desktop platform for cytogenetic analysis assisted by \textit{Deep Learning}. The experiments on the CRCN-NE dataset demonstrated that YOLOv11 achieves 99.40\% mAP@50, confirming the robustness of single-stage detectors for this task. By integrating this performance with a complete GUI featuring \textit{benchmarking} and interactive annotation modules, Aycromo overcomes the accessibility barrier documented in the literature, where most tools lack a graphical interface or operate in a private domain, and reduces the slide analysis time.

As future work, we plan to include support for the automatic classification of the 23 homologous pairs, the addition of new state-of-the-art models, and the formal usability evaluation with specialists in a real clinical environment.

\bibliographystyle{sbc}
\bibliography{sbc-template}

\end{document}